\documentclass[]{llncs}
\usepackage{graphicx}
\usepackage{booktabs}
\usepackage{xspace}
\usepackage{amssymb}
\usepackage{subfig}
\usepackage{algorithm}
\usepackage{algpseudocode}
\usepackage{color}
\usepackage{amsmath}
\usepackage{multirow}
\usepackage{wrapfig}
\usepackage{lipsum}

\makeatletter
\DeclareRobustCommand\onedot{\futurelet\@let@token\@onedot}
\def\@onedot{\ifx\@let@token.\else.\null\fi\xspace}

\def\eg{\emph{e.g}\onedot} 
\def\ie{\emph{i.e}\onedot}

\def\etal{\emph{et al}\onedot}
\makeatother

\begin{document}

\title{Difficulty-aware Meta-learning \\for Rare Disease Diagnosis}
\author{Xiaomeng Li\inst{1,2}  \and 
	Lequan Yu\inst{1,2} \and
	Yueming Jin\inst{1} \and 
	Chi-Wing Fu\inst{1}\and \\
	Lei Xing\inst{2} \and 
	Pheng-Ann Heng\inst{1,3}
}

\institute{
	Dept. of Computer Science and Engineering, The Chinese University of Hong Kong\\
	\and Dept. of Radiation Oncology, Stanford University, United States \\
	\and Shenzhen Key Laboratory of Virtual Reality and Human Interaction Technology, Shenzhen Institutes of Advanced Technology, Chinese Academy of Sciences, China}

\newcommand{\TODO}[1]{{\color{red}{[TODO: #1]}}}
\newcommand{\phil}[1]{{\color{green}{#1}}}
\newcommand{\ylq}[1]{{\color{blue}{[LQ: #1]}}}
\maketitle

\begin{abstract}
	Rare diseases have extremely low-data regimes, unlike common diseases with large amount of available labeled data.
	Hence, to train a neural network to classify rare diseases with a few per-class data samples is very challenging, and so far, catches very little attention.
	In this paper, we present a difficulty-aware meta-learning method to address rare disease classifications and demonstrate its capability to classify dermoscopy images.
	Our key approach is to first train and construct a meta-learning model from data of common diseases, then  adapt the model to perform rare disease classification.
	To achieve this, we develop the difficulty-aware meta-learning method that dynamically monitors the importance of
	learning tasks during the meta-optimization stage. 
	To evaluate our method, we use the recent ISIC 2018 skin lesion classification dataset, and show that with only five samples per class, our model can quickly adapt to classify unseen classes by a high AUC of 83.3\%.
	Also, we evaluated several rare disease classification results in the public Dermofit Image Library to demonstrate the potential of our method for real clinical practice.	
\end{abstract}

\section{Introduction}
Deep learning methods have become the de facto standard for many medical imaging analysis tasks, \eg, anatomical structure segmentation and computer-aided disease diagnosis. 
One reason of the success is due to 
a large amount of labeled data to support the network training. 
Yet, there are about 7,000 known rare diseases~\cite{jia2018rdad} that typically catch little attention and the data is difficult to obtain. 
These conditions collectively affect about 400 million people worldwide~\cite{khoury2016rare} and were generally neglected by the medical imaging community. 
Taking the retinal diseases as an example, the glaucoma, diabetic retinopathy, age-related macular degeneration, and retinopathy of prematurity are relatively common in the clinical practice.
Whereas, other retinal diseases, such as fundus pulverulentus and fundus albipunctatus are rare~\cite{skorczyk2015fundus}.
Generally, it is difficult to collect data for these rare diseases and obtain annotations from experienced physicians.
This phenomenon raises the following question:~
\emph{given the extremely low-data regime of rare diseases, can we transfer the inherent knowledge learned from the common diseases to support the automated rare disease diagnosis?}

A vanilla solution is transfer learning,~\ie, fune-tuning, a widely-used and effective approach
, where the model is usually transferred from one dataset to another smaller one. 
However, fine-tuning a model on an extremely low-data regime will severely overfit to the few given data, \ie, \emph{less than five samples per class}; see the training curves shown in Fig.~\ref{fig:baseline} in the experiments. 
Another possible solution to alleviate overfitting problem is to use a pretrained network to extract features, and then utilize another classifiers, \eg, support vector machine (SVM) and k-nearest neighbors (KNN), to perform the classification.
However, a series of experiments in section 3 show that these methods have limited capacity in the classification of rare diseases.
Recently, Zhao~\etal~\cite{zhao2019data} and Mondal~\etal~\cite{mondal2018few} tackle the low-data regime related issues, \ie, one-shot or few-shot segmentation problems, in the medical image domain.
However, both  works relies on the large amount of unlabeled images, which are not appropriate to handle rare disease diagnosis.
Maicas~\etal~\cite{maicas2018training} present a meta-learning method to learn a good initialization on a series of tasks, which can be used to pre-train medical image analysis models.
In this regard, developing effective techniques for  rare disease diagnosis in low-data regime is of vital importance.

Meta-learning techniques, or learning to learn, is the science of systematically observing how an  algorithm performs on a wide range of learning tasks, and then learning from this experience, \ie, meta-data, to learn new tasks much faster.
In general, meta-learning updates a network by equally treating (averaging) the gradient directions of different randomly-sampled tasks. So, the meta-learning process often stops at a stage, at which ``easy tasks'' are well-learned and ``difficulty tasks'' are still being misclassified. This hinders the meta-learning and affects the results on rare disease classification, in which the rare disease samples are unseen in the training. This observation motivates us to consider a more effective meta-learning optimization.
To this end, we propose a novel difficulty-aware meta-learning (DAML) method.
Our method first train a meta-classifier on a series of related tasks (\eg, common disease classification), instead of an individual single task, such that the transferable internal representations with the gradient-based learning rule can make rapid progress on the new tasks (\eg, rare disease classification). 
More importantly, we discover that the contribution of each task sample to the meta-objective is various.
To better optimize the meta-classifier, a dynamic modulating function over the learning tasks is formulated, where the function automatically down-weights the well-learned tasks and rapidly focuses on the hard tasks. 
Our method is evaluated on the ISIC 2018 Skin Lesion Dataset~\cite{codella2018skin,li2018deeply}, where the data are annotated with seven lesion categories. 
Only training on the four skin lesion classes, our method achieves a promising result on classifying other three unseen classes, with an AUC of 83.3\% under five samples setting. 
We also validate our method on several rare disease classification tasks using the public Dermofit Image Library\footnote{https://licensing.eri.ed.ac.uk/i/software/dermofit-image-library.html} and achieve a high AUC of 82.67\% under the five samples setting, demonstrating the potential of our method for real clinical practice.

\begin{figure}[t]
	\centering
	\includegraphics[width=0.9\textwidth]{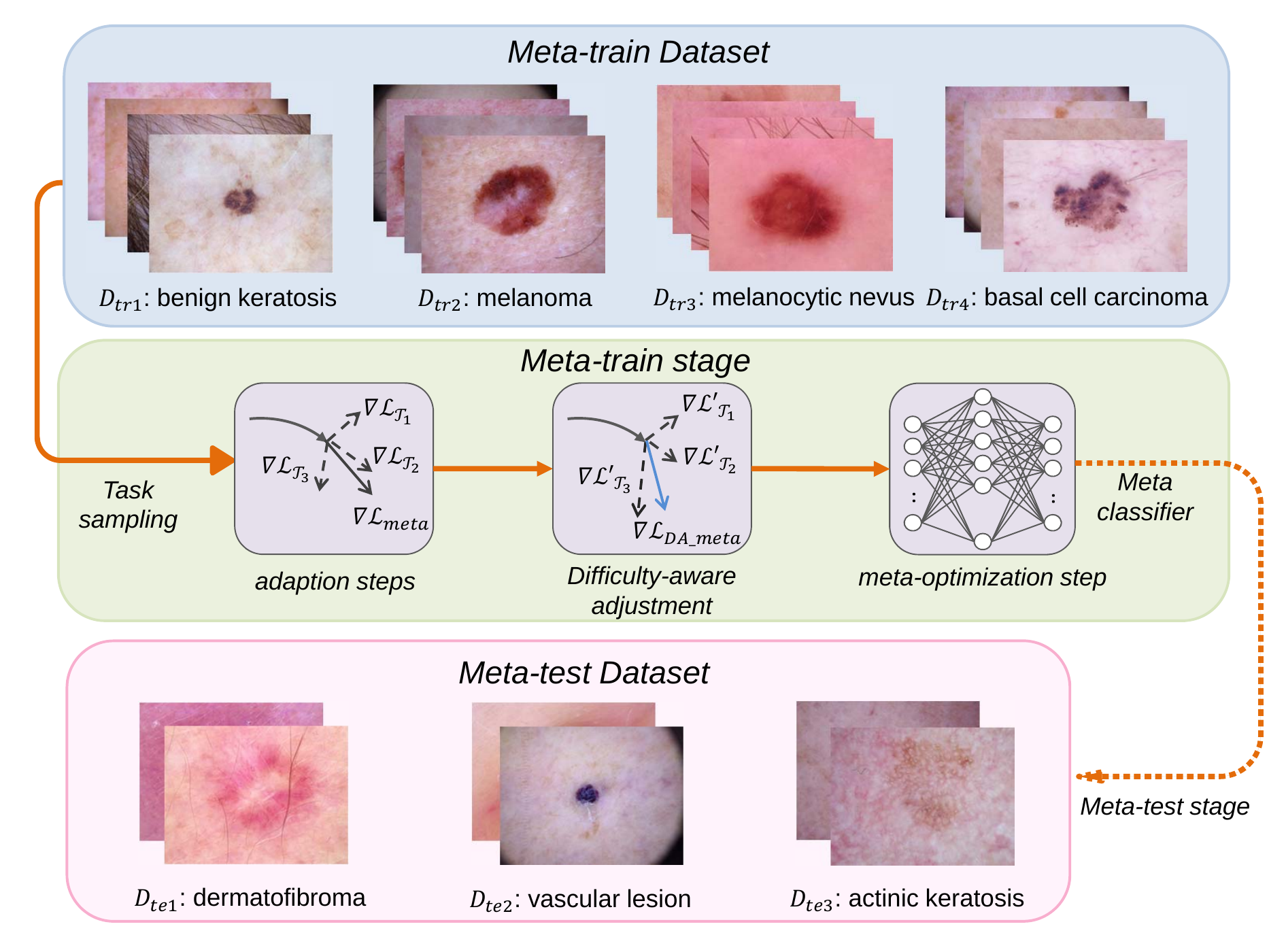}
	\caption{The pipeline of our proposed difficulty-aware meta-learning (DAML) system. The meta-classifier (neural network) is explicitly trained on the meta-train dataset, such that given new tasks with only a few samples, the meta-classifier can rapidly adapt to the new tasks with a high accuracy. Our novel difficulty-aware meta-optimization scheme can dynamically down-weight the contribution of easy tasks and focus more to learn from hard tasks.}
	\label{fig:framework}
	
\end{figure}

\section{Method}

We aim to train a neural network using meta-train data (common diseases), such that given new tasks associated with few data samples (meta-test data for rare diseases), we can quickly adapt the network model via a few steps of gradient descent to handle the new tasks; see Fig.~\ref{alg:Framwork} for the pipeline.

\subsection{Problem Setting}
We employ the ISIC 2018 Skin Lesion Analysis Towards Melanoma Detection Dataset~\cite{milton2019automated,li2020transformation}, which has a total of 10,015 skin lesion images from seven skin diseases, including melanocytic nevus (6705), melanoma (1113),  benign keratosis (1099), basal cell carcinoma (514), actinic keratosis (327), vascular lesion (142) and  dermatofibroma (115). 
We simulate the problem by utilizing the four classes with largest amount of cases as common diseases (\ie, meta-train dataset $D_{tr}$) and the left three classes as the rare diseases (\ie, meta-test dataset $D_{te}$).


Task instance $\mathcal{T}_i$ is randomly sampled from distribution over tasks $p(\mathcal{T})$ and $D_{tr}, D_{te} \in p(\mathcal{T})$.
During meta-train stage,
learning task $\mathcal{T}_i$ are binary classification tasks and each task  consists of two random classes with $k$ samples per class in $D_{tr}$.
During the meta-test stage, each test task instance is sampled from $D_{te}$.

\subsection{Difficulty-aware Meta-learning Framework}
Current model-agnostic meta-earning learns the meta-classifier according to the averaged evaluation of tasks sampled from $p(\mathcal{T})$~\cite{finn2017model}. 
However, this meta-learning process is easily dominated by well-learned tasks.
To improve the effectiveness and emphasize on difficult tasks in the meta-training stage, we propose the difficulty-aware meta-learning method. 
The main framework and meta-training procedure are described in Fig.~\ref{fig:framework} and Algorithm~\ref{alg1}.
We \textbf{meta-train} a base model parameters $\phi$ on a series of learning tasks in $D_{tr}$.
First, task instance $\mathcal{T}_i$ is randomly sampled from $p(\mathcal{T})$ (line 3 in Algorithm~\ref{alg1}).  
As mentioned above, the learning task $\mathcal{T}_i$ is a binary classification task that consists of two random classes with $k$ samples per class in $D_{tr}$.
The ``adaptation steps'' takes $\phi$ as input and returns parameters $\phi_i'$ adapted specifically for task instance $\mathcal{T}_i$ by using gradient descent iteratively, with the corresponding cross entropy loss function $\mathcal{L}_{\mathcal{T}_i}$ for $\mathcal{T}_i$ (lines 5-8). 
The cross-entropy loss for $\mathcal{T}_i$ and the gradient descent are defined in Eq.~(\ref{eq0}) and Eq.~(\ref{eq00}).

\begin{equation}
\begin{split}
\mathcal{L}_{\mathcal{T}_i} \left (f{_{\phi_i}}  \right) = -\sum_{x_j, y_j \sim \mathcal{T}_i} y_j \ {\rm log}(f_{\phi_i}(x_j)) + (1- y_j) \ {\rm log}(1- f_{\phi_i}(x_j)),
\end{split}
\label{eq0}
\end{equation}

\begin{equation}
{\phi}'_{i} \leftarrow  \phi_{i} - \gamma \triangledown _\phi \mathcal{L}_{\mathcal{T}_i}(f_{\phi_i}) ,
\label{eq00}
\end{equation}
where $\gamma$ is the inner loop adaptation learning rate and $\phi'_i$ is the adapted model parameters for task $\mathcal{T}_i$.

\begin{algorithm}[htbp]  
	\caption{Meta Learning Algorithm}  
	\label{alg:Framwork} 
	\small 	
	\begin{algorithmic}[1]  
		\Require $D_{tr}$:  Meta-train dataset
		\Require $\alpha$, $\gamma$: meta learning rate, inner-loop adaptation learning rate 
		\State Randomly initialize network weight $\phi $
		\While {not converged} 
		\State Sample batch of tasks from $D_{tr}$
		\For {task $\mathcal{T}_i$ in batch} 
		
		\For {number of adaptation steps}
		\State	Evaluate  $  \mathcal{L}_{\mathcal{T}_i} \left (f{_{\phi_i}}  \right) $ with respect to $k$ samples using Eq.~(\ref{eq0}).
		
		\State Compute gradient descent for $\mathcal{T}_i$ using Eq.~(\ref{eq00}).
		\EndFor
		\State 
		Evaluate $  \mathcal{L}_{DA\_meta} \left (f{_{\phi_i}}  \right) $  using Eq.~\eqref{eq:3}. 
		\EndFor 
		\State Update network weight $\phi$ using Eq.~(\ref{eq:4}). 
		\EndWhile
		\\ \Return $\phi$;  	
	\end{algorithmic}  
	\label{alg1}
\end{algorithm}

\begin{wrapfigure}{r}{0.5\textwidth}
	\includegraphics[width=0.5\textwidth]{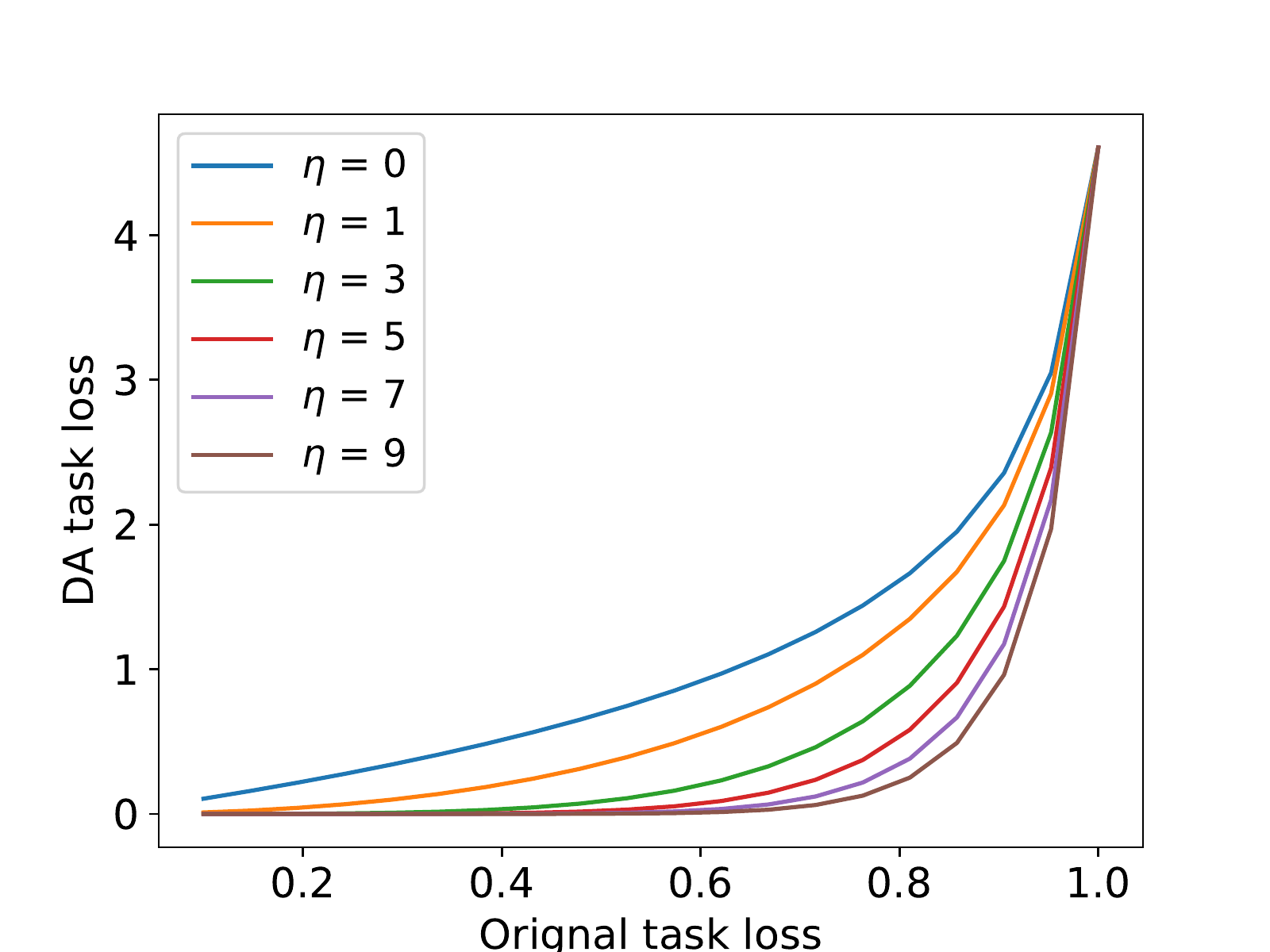}
	\caption{Visualization of Eq.~(\ref{eq:3}). The original task loss could be infinite. For clear visualization, we show the original task loss within [0, 1]. } \label{fig:function}
\end{wrapfigure}

After the ``adaptation steps'', 
the model parameters are trained by optimizing for the performance respect to tasks adapted in the ``adaptation steps''.
Concretely, a dynamically scaled cross-entropy loss \textbf{over the learning tasks} is formulated, where it automatically down-weights the easy tasks and focuses on hard tasks; as shown in Fig~\ref{fig:function}.
Formally, the difficulty-aware meta optimization function is defined as
\begin{equation}
\mathcal{L}_{DA\_meta} =  \sum_{\mathcal{T}_i \sim p(\mathcal{T})} {-\mathcal{L}_{\mathcal{T}_{i}}}^ {\eta} \ {\rm log  (\max \ }(\epsilon, 1-\mathcal{L}_{\mathcal{T}_{i}})),
\label{eq:3}
\end{equation}

\begin{equation}
\phi \leftarrow \phi - \alpha \triangledown_\phi \sum_{\mathcal{T}_i \sim p(\mathcal{T})}^{} \mathcal{L}_{\mathcal{T}_{DA\_meta}},
\label{eq:4}
\end{equation}
where $\mathcal{L}_{\mathcal{T}_i}$ is the original cross entropy loss for task $\mathcal{T}_i$, $\eta$ is a scaling factor 
and $\epsilon$ is a smallest positive integer satisfying $ \max \ (\epsilon, 1 - \mathcal{L}_{\mathcal{T}_i} )>0$. 
Then the meta-classifier is updated by performing Eq.~\eqref{eq:4}.
The whole meta-training procedure can be found in Algorithm~\ref{alg1}.

Intuitively, our difficulty-aware meta-loss dynamically reduces the contribution from easy tasks and focuses on the hard tasks, which in turn increases the importance of optimizing the misclassified tasks.
For example, with $\eta$ = 3, a task with cross-entropy loss 0.9 would have higher loss in the meta-optimization stage, while a task with lower loss would be given less importance. 
We analyze the effect of different values for hyperparameter $\eta$ in the following experiments. 

\setlength\intextsep{-1pt}
\noindent

\subsection{Meta-training Details} 
We employed the 4 conv blocks as the backbone architecture~\cite{sung2018learning}, which has 4 modules with a 3x3 conv and 64 filters, followed by a BN, a ReLU, and a 2x2 max-pooling.
We used Adam optimizer with a meta-learning rate of 0.001 and divide by 10 for every 150 epochs.
We totally trained 3000 iterations and the adopted the difficulty-aware optimization at around 1500 iterations.
The batch size is 4, consisting of 4 tasks sampled from meta-train dataset.
Each task consists of randomly $k$ samples from 2 classes.
We query 15 images from each of two classes to adapt parameters for $\mathcal{T}_i$, as the same protocol in~\cite{sung2018learning,finn2017model,li2020revisiting}.
During meta-test stage, the inference is performed by randomly sampling $k$ samples from 2 classes from meta-test dataset, \ie, $D_{te}$.
The final report results is the averaged AUC over 30 runs.

\section{Experiments and Results}
We conduct experiments on ISIC 2018 skin lesion classification dataset\footnote{https://challenge2018.isic-archive.com/task3/}.
We first compare our method with some strong baselines, \ie, fine-tuning, feature extraction+classifiers, to validate the effectiveness of our method for classification in  the extremely low-data regime.
Then, we compare with other related methods to show the effectiveness of our novel difficulty-aware meta-learning framework.
Next, we analyze the improvement of our proposed difficulty-aware meta-optimization loss, the effect of different scaling factors, as well as the importance of network architecture and data augmentation.

\begin{figure}
	\centering
	\subfloat[Training loss of Fine-tuning method]{{\includegraphics[width=5.5cm]{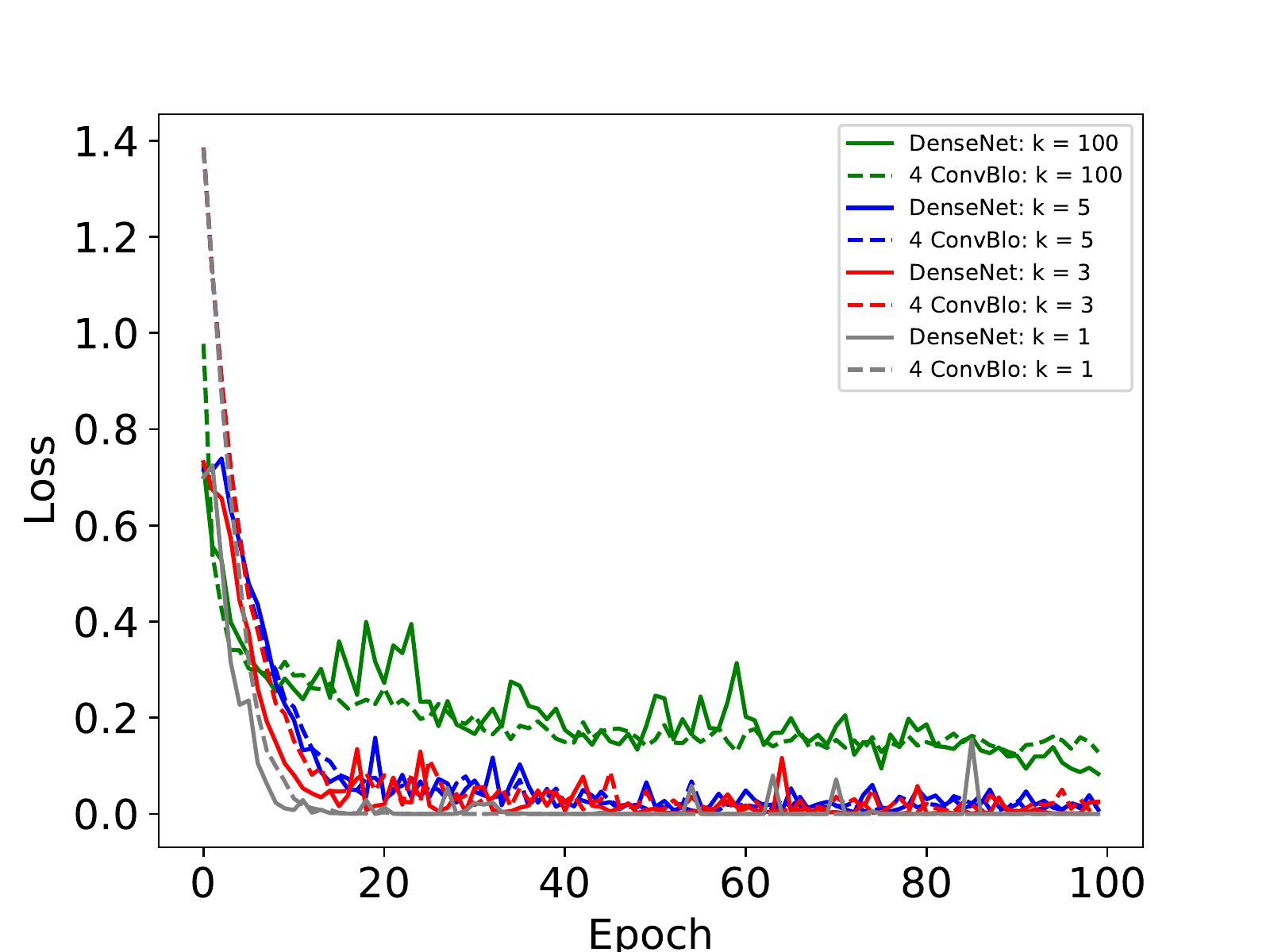}}}%
	\qquad
	\subfloat[AUC result of Fine-tuning method ]{{\includegraphics[width=5.5cm]{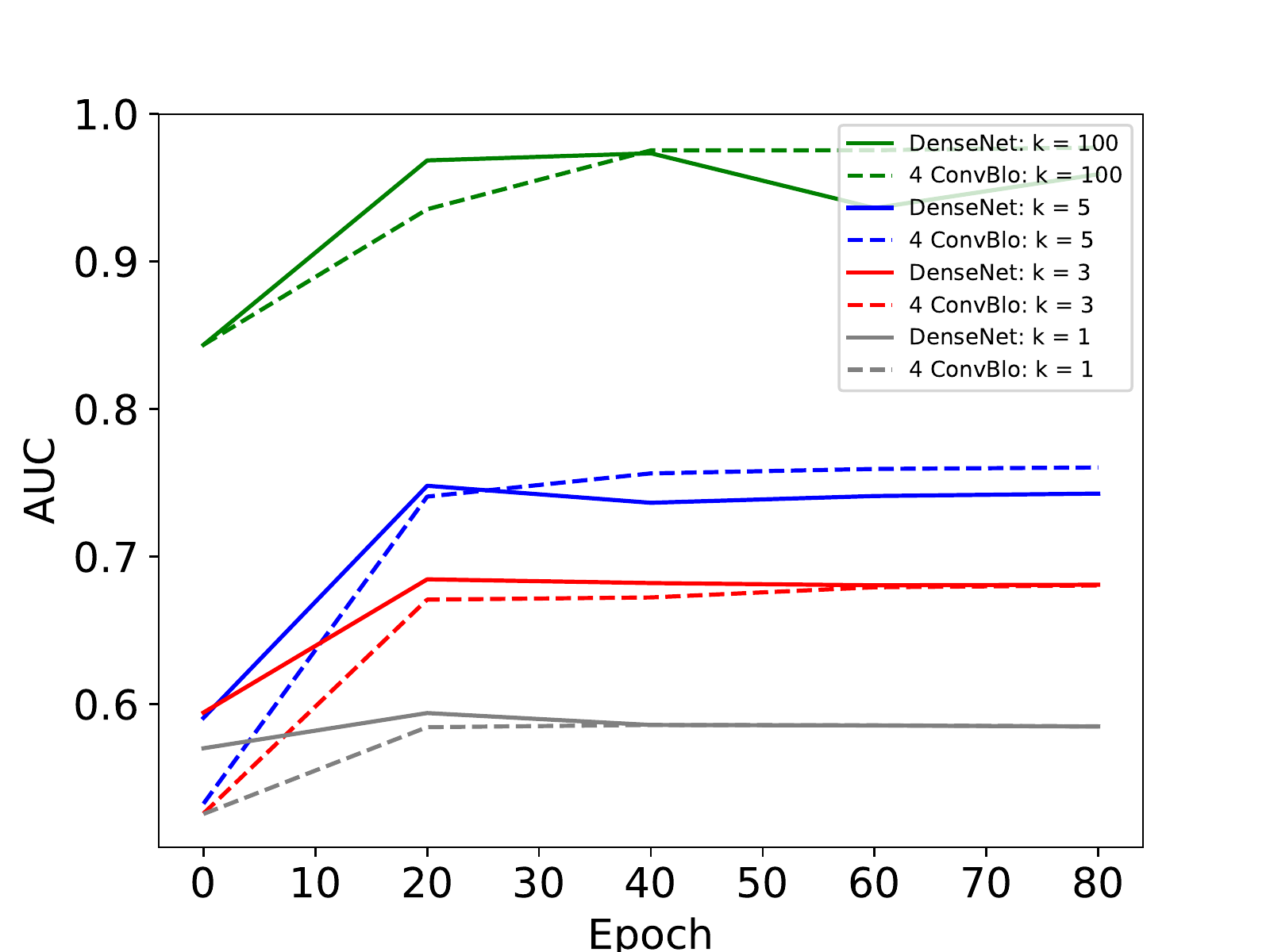} }}%
	\caption{The training curve (a) and test AUC (b) when performing fine-tuning on meta-test dataset. 
	}
	\label{fig:baseline}
	
\end{figure}

\subsubsection{Comparison with Strong Baselines}
We first show the results of the standard fine-tuning method on rare disease classification for comparison. 
We report the performance of fine-tuning with two different network architectures: the DenseNet~\cite{huang2017densely} and 4 Conv Blocks~\cite{sung2018learning} in Table~\ref{tab:result}, where the former one is the state-of-the-art classification network and the latter one is the most commonly used architecture in few-shot learning setting in computer vision. 
Note that we employ the data augmentation technique on both the pre-training and fine-tuning stages, including random scaling, rotation, and mirror flipping.
Fig.~\ref{fig:baseline} shows the training loss curve and the test AUC performance when performing fine-tuning on few given samples. Each plot is the average over 30 runs. 
It is observed that when $k$ is large, the fine-tuning could perform well and achieves the AUC at around 97\%. 
However, with a smaller $k$, the model converges rapidly but the AUC performance reduces devastatingly.
For example, the 4 Conv Blocks only achieves 58.49\% AUC with $k=1$.
We also employ the pre-trained DenseNet as feature extractor and utilize another classifiers, \eg, SVM and KNN, to coduct rare disease classification.
As shown in Table~\ref{tab:result}, the performance of these methods is inferior, indicating that these methods have limited capacity in tackling the classification task with just a few samples from unseen class.
In other aspect, our method achieves 83.30\% AUC with $k=5$ and 79.60\% AUC with $k=3$, demonstrating the effectiveness of our method for disease classification in the extremely low-data regime. 
\begin{table}[!t]
\centering
\caption{The AUC performance of different methods on skin lesion dataset. Each result is averaged over 30 runs.}
\resizebox{0.95\columnwidth}{!}{
	\begin{tabular}{c | c|| c| c|| c |c ||c|c}
		\toprule[1.5pt] 
		& backbone & Sample \#& AUC & Sample \#& AUC & Sample \#& AUC \tabularnewline \hline			
		ConvFeature + KNN & DenseNet & 1  & 50.00\% & 3 & 56.07\% & 5  & 62.69\% \tabularnewline 
		ConvFeature + SVM  & DenseNet  & 1  &  61.46\% & 3 & 61.68\% & 5  & 67.44\% \tabularnewline 
		
		Finetune + Aug &  DenseNet & 1  & 58.57\% & 3 & 68.05\% & 5  & 73.65\% \tabularnewline 
		Finetune + Aug &  4 Conv Blocks & 1  & 58.49\%& 3 & 68.13\% & 5  & 75.90\%  \tabularnewline \hline
		Relation Net~\cite{sung2018learning} 
		& 4 Conv Blocks & 1 & 59.97\% & 3 & 62.87\%  & 5 & 72.40\%
		\tabularnewline 	
		
		MAML ~\cite{finn2017model}
		&   4 Conv Blocks & 1 & 63.77\% & 3 & 77.98\% & 5 & 81.20\%  \tabularnewline 
		
		Task sampling~\cite{maicas2018training} & 4 Conv Blocks & 1 & 64.21\% & 3  & 78.40\% & 5  & 82.05\% \tabularnewline

		DAML (ours)    &  4 Conv Blocks & 1  & \textbf{67.33\%} & 3 & \textbf{79.60\%} & 5  & \textbf{83.30\%} \tabularnewline 
		\bottomrule[1.5pt]
	\end{tabular}}
	\label{tab:result}
\end{table}

\subsubsection{Comparison with Other Methods}
We also report the performance of the widely used few shot learning approaches, Relation Net~\cite{sung2018learning}, MAML~\cite{finn2017model} and Task sample~\cite{maicas2018training} for the rare disease classification task in Table~\ref{tab:result}.
Our method surpasses these three approaches, especially the Relation Net.
The reason may be that our meta-train dataset only has four classes, limiting the representation capability of the feature extraction in the metric-based approach.  
It is worth noting that the meta-learning based approaches excels all the baseline methods and metric-based method, demonstrating the promising results of meta-learning approach for rare disease classification. 
Overall, our method further improves the original MAML method, which demonstrates the effectiveness of our difficulty-aware meta optimization procedure. 
Maicas~\etal~\cite{maicas2018training} proposes a meta-learning method that addresses the task sampling issue. From Table~\ref{tab:result}, we can see our method achieves better results than task sampling method~\cite{maicas2018training} under all sample settings.

\subsubsection{Ablation study of Our Method}
We provide detailed analysis on the effects of our difficulty-aware meta-optimization loss, network architectures and data augmentation under 5 samples setting, as shown in Table~\ref{table:differentdata}.
First, we analyze the effects of $\eta$  in our difficulty-aware loss. 
We found that our method can obviously improve the overall performance of the meta-learning and the performance is best when $\eta = 5$. 
We then explore the importance of network architecture. ``conv blocks + 64f + Aug'' refers to architectures consisting of 4 conv blocks with 64 feature maps with heavy data augmentation. More complicated networks, \ie, residual blocks, would severely lead to the overfitting problem; as the comparison shown in Table~\ref{table:differentdata}.
Moreover, the heavy data augmentation, \ie, random rotation, flipping, scaling with crop, can only slightly improve the AUC results from 79.3\% to 81.2\%.

\begin{figure}[!t] 
\centering
\captionsetup{type=table} 
\caption {Ablation study results under 5 samples setting.} 
{\begin{tabular}{c|c}
		\toprule[1.5pt] 
		Experiments setting     & AUC  \tabularnewline \hline	
		ResBlocks + 32f + no Aug    &  75.8\%   \tabularnewline 	
		%
		ConvBlocks + 64f + no Aug    & 79.3\% \tabularnewline 
		ConvBlocks + 64f + Aug    & 81.2\%  \tabularnewline 
		\hline 	
		ConvBlocks + 64f + Aug, $\eta=1$   & 81.7\%   \tabularnewline 
		ConvBlocks + 64f + Aug, $\eta=3$  & 82.4\%   \tabularnewline 
		ConvBlocks + 64f + Aug, $\eta=5$  & \textbf{83.3\%}   \tabularnewline 
		ConvBlocks + 64f + Aug, $\eta=7$  & 83.1\%  \tabularnewline 
		\bottomrule[1.5pt]	 
		
	\end{tabular}}
	
	\label{table:differentdata}
	
\end{figure}

\begin{table}[t]
	\centering
	\caption{Results of our method and other method for rare disease classification on the Public Dermofit Image Library.}
	{\begin{tabular}{cccccccc}
			\toprule[1.5pt] 
			& backbone & Sample & AUC & Sample & AUC & Sample & AUC \tabularnewline \hline					
			MAML ~\cite{finn2017model}
			&   4 Conv Blocks & 1 & 63.00\% & 3 & 74.03\% & 5 & 80.70\%  \tabularnewline 
			
			Task sampling~\cite{maicas2018training} & 4 Conv Blocks & 1 & 63.20\%  & 3  & 76.10\%  & 5  &  81.80\% \tabularnewline 
			DAML (ours)    &  4 Conv Blocks & 1  & \textbf{63.33\%} & 3 & \textbf{77.15\%} & 5  & \textbf{82.76\%} \tabularnewline 
			\bottomrule[1.5pt]	 
		\end{tabular}}
		\label{tab:raredisease}
		
	\end{table}

	\subsubsection{Validation on Real Clinical Data}
	We validated our method for rare diseases classification in the real clinical data from public Dermofit Image Library\footnote{https://licensing.eri.ed.ac.uk/i/software/dermofit-image-library.html}. 
	The disease classes we employed are \textit{squamous cell carcinoma}, \textit{haemangioma}, and \textit{pyogenic granuloma}.
	As shown in Table~\ref{tab:raredisease}, our method outperforms other related meta-learning methods.  
	Our method achieves an average AUC of 82.67\% under five samples setting, demonstrating the potential usage of our method for real clinical applications.
	
	\subsubsection{Discussions on Other Applications}
	Our method has the potential to be applied to other related applications such as federated learning.  
	Medical data usually has privacy regulations, hence, it is often infeasible
	to collect and share patient data in a centralized data lake. 
	This issue poses challenges for training deep convolutional networks, which often require large numbers of diverse training examples. 
	Federated learning provides a solution, which allows collaborative and decentralized training of neural networks without sharing the patient data.
	For example, Li~\etal~\cite{li2019privacy} implemented and evaluated practical federated learning systems for brain tumor segmentation.
	Our method is also feasible to be tested for federated learning since our method only accesses training data (common dataset) during the training stage and can be fast adapted according to the test data (private dataset) during the inference time.   
	Exploring the applications of our method in federated learning would be a future work of this paper. 
	

\section{Conclusion}
In this paper, we propose a novel difficulty-aware meta-learning method to tackle the extremely low-data regime problem, \ie, rare disease classification.
Our difficulty-aware meta learning approach optimizes the meta-optimization stage by down-weighting the well classified tasks and emphasizing on hard tasks.  
Extensive experiments demonstrated the superiority our method. Our results excels other strong baselines as well as other related methods.
The clinical rare skin disease cases from Dermofit Image Library also validated our method for piratical usage.
\section*{Acknowledgements} 
The work described in the paper was supported in parts by the following grants from
Key-Area Research and Development Program of Guangdong Province, China (2020B010165004), 
Hong Kong Innovation and Technology Fund (Project No. ITS/311/18FP $\&$ ITS/426/17FP) 
and National Natural Science Foundation of China (Project No. U1813204).

\bibliographystyle{splncs04}
\tiny{\bibliography{refs}}
\end{document}